\documentclass[11pt,a4paper]{article}
\usepackage[hyperref]{emnlp2020}
\usepackage{times}
\usepackage{latexsym}

\usepackage{microtype}

\aclfinalcopy %
\def\aclpaperid{286} %

\usepackage{booktabs}
\usepackage{amsmath}
\usepackage{graphicx}
\usepackage{amsfonts,amssymb}
\usepackage{xspace}
\usepackage{multirow}
\usepackage[normalem]{ulem}

\newcommand{\ie}{\textit{i.e.}} %
\newcommand{\eg}{\textit{e.g.}} %
\newcommand{\start}[1]{\vspace{1.1mm}\noindent{{\bf #1}.}}

\DeclareMathOperator*{\argmax}{arg\,max\,}

\newcommand{\upv}{\vspace{-.0cm}}
\newcommand{\downv}{\vspace{-.0cm}}
\newcommand{\upve}{\vspace{-.0cm}}
\newcommand{\downve}{\vspace{-.0cm}}

\newcommand{\ours}{\textsc{RL-MMR}\xspace}
\newcommand{\oursHard}{$\text{RL-MMR}_{\textsc{Hard-Cut}}$\xspace}
\newcommand{\oursHardComb}{$\text{RL-MMR}_{\textsc{Hard-Comb}}$\xspace}
\newcommand{\oursSoftComb}{$\text{RL-MMR}_{\textsc{Soft-Comb}}$\xspace}
\newcommand{\oursSoftAttn}{$\text{RL-MMR}_{\textsc{Soft-Attn}}$\xspace}

\newcommand{\Hard}{\textsc{Hard-Cut}\xspace}
\newcommand{\HardComb}{\textsc{Hard-Comb}\xspace}
\newcommand{\SoftComb}{\textsc{Soft-Comb}\xspace}
\newcommand{\SoftAttn}{\textsc{Soft-Attn}\xspace}

\usepackage{amsmath,amsfonts,bm}

\newcommand{\figleft}{{\em (Left)}}

\newcommand{\figright}{{\em (Right)}}

\def\eqref#1{equation~\ref{#1}}

\def\1{\bm{1}}

\def\rmR{{\mathbf{R}}}
\def\rmS{{\mathbf{S}}}

\def\va{{\bm{a}}}

\def\vg{{\bm{g}}}
\def\vh{{\bm{h}}}

\def\vm{{\bm{m}}}

\def\vv{{\bm{v}}}

\def\vz{{\bm{z}}}

\def\mA{{\bm{A}}}

\def\mE{{\bm{E}}}

\def\mM{{\bm{M}}}

\def\mW{{\bm{W}}}

\DeclareMathAlphabet{\mathsfit}{\encodingdefault}{\sfdefault}{m}{sl}
\SetMathAlphabet{\mathsfit}{bold}{\encodingdefault}{\sfdefault}{bx}{n}

\def\gD{{\mathcal{D}}}

\def\gR{{\mathcal{R}}}

\newcommand{\softmax}{\mathrm{softmax}}

\definecolor{gred}{RGB}{219,68,55}
\definecolor{gblue}{RGB}{66,133,244}
\definecolor{gyellow}{RGB}{244,180,0}
\definecolor{ggreen}{RGB}{15,157,88}
\definecolor{ggrey}{RGB}{115,115,115}

\newcommand{\colorR}[1]{\textcolor{gred}{\textbf{#1}}}
\newcommand{\colorG}[1]{\textcolor{ggreen}{\textbf{#1}}}
\newcommand{\colorB}[1]{\textcolor{gblue}{\textbf{#1}}}

\title{Multi-document Summarization with Maximal Marginal Relevance-guided Reinforcement Learning}

\author{Yuning Mao$^{1}$, Yanru Qu$^{1}$, Yiqing Xie$^{1}$, Xiang Ren$^2$, Jiawei Han$^1$ \\
$^1$University of Illinois at Urbana-Champaign, IL, USA \\
$^2$University of Southern California, CA, USA\\
$^1$\{yuningm2, yanruqu2, xyiqing2, hanj\}@illinois.edu $\quad$ $^2$xiangren@usc.edu
}
\date{}

\begin{document}
\maketitle

\begin{abstract}
While neural sequence learning methods have made significant progress in single-document summarization (SDS), they produce unsatisfactory results on multi-document summarization (MDS).
We observe two major challenges when adapting SDS advances to MDS: (1) MDS involves larger search space and yet more limited training data, setting obstacles for neural methods to learn adequate representations; (2) MDS needs to resolve higher information redundancy among the source documents, which SDS methods are less effective to handle.
To close the gap, we present \ours, Maximal Margin Relevance-guided Reinforcement Learning for MDS, which unifies advanced neural SDS methods and statistical measures used in classical MDS. 
\ours casts MMR guidance on fewer promising candidates, which restrains the search space and thus leads to better representation learning. Additionally, the explicit redundancy measure in MMR helps the neural representation of the summary to better capture redundancy.
Extensive experiments demonstrate that \ours achieves state-of-the-art performance on benchmark MDS datasets.
In particular, we show the benefits of incorporating MMR into end-to-end learning when adapting SDS to MDS in terms of both learning effectiveness and efficiency.\footnote{Code can be found at \url{https://github.com/morningmoni/RL-MMR}.}

\end{abstract}

\section{Introduction}
Text summarization aims to produce condensed summaries covering salient and non-redundant information in the source documents.
Recent studies on single-document summarization (SDS) benefit from the advances in neural sequence learning ~\cite{nallapati-etal-2016-abstractive,see-etal-2017-get, chen-bansal-2018-fast, narayan-etal-2018-ranking} as well as pre-trained language models \cite{liu-lapata-2019-text,lewis2019bart,zhang2019pegasus} and make great progress. 
However, in multi-document summarization (MDS) tasks, neural models are still facing challenges and often underperform classical statistical methods built upon handcrafted features~\cite{kulesza2012determinantal}.

We observe two major challenges when adapting advanced neural SDS methods to MDS: 
(1) \textbf{Large search space}. 
MDS aims at producing summaries from multiple source documents, which
exceeds the capacity of neural SDS models \cite{see-etal-2017-get} and sets learning obstacles for adequate representations, especially considering that MDS labeled data is more limited.
For example, there are 287K training samples (687 words on average) on the CNN/Daily Mail SDS dataset~\cite{nallapati-etal-2016-abstractive} and only 30 on the DUC 2003 MDS dataset (6,831 words).
(2) \textbf{High redundancy}.
In MDS, the same statement or even sentence can spread across different documents.
Although SDS models adopt attention mechanisms as implicit measures to reduce redundancy~\cite{chen-bansal-2018-fast}, they fail to handle the much higher redundancy of MDS effectively (Sec. \ref{sec:exp_mmr_guidance}).

There have been attempts to solve the aforementioned challenges in MDS.
Regarding the \textbf{large search space},
prior studies \cite{lebanoff-etal-2018-adapting, zhang-etal-2018-adapting} perform sentence filtering using a sentence ranker and only take top-ranked $K$ sentences.
However, such a hard cutoff of the search space makes these approaches insufficient in the exploration of the (already scarce) labeled data and limited by the ranker since most sentences are discarded,\footnote{$K$ is set to 7 in~\citet{lebanoff-etal-2018-adapting} and 15 in~\citet{zhang-etal-2018-adapting}. One document set in DUC 2004~\cite{paul2004introduction}, for example, averages 265.4 sentences.} albeit the discarded sentences are important and could have been favored.
As a result, although these studies perform better than directly applying their base SDS models \cite{see-etal-2017-get, tan-etal-2017-abstractive} to MDS, 
they do not outperform state-of-the-art MDS methods~\cite{gillick-favre-2009-scalable, kulesza2012determinantal}. 
Regarding the \textbf{high redundancy}, 
various redundancy measures have been proposed, including heuristic post-processing such as counting new bi-grams \cite{cao-etal-2016-attsum} and cosine similarity~\cite{hong-etal-2014-repository}, or dynamic scoring that compares each source sentence with the current summary like Maximal Marginal Relevance (MMR)~\cite{carbonell1998use}.   Nevertheless, these methods still use lexical features without semantic representation learning. One extension~\cite{cho-etal-2019-improving} of these studies uses capsule networks~\cite{hinton2018matrix} to improve redundancy measures. However, its capsule networks are pre-trained on SDS and fixed as feature inputs of classical methods  without end-to-end representation learning.

In this paper, we present a deep RL framework, MMR-guided Reinforcement Learning (\ours) for MDS, which unifies advances in SDS and one classical MDS approach, MMR~\cite{carbonell1998use} through end-to-end learning.
\ours addresses the MDS challenges as follows: (1) \ours overcomes the \textbf{large search space} through \textit{soft attention}. Compared to hard cutoff, our soft attention favors top-ranked candidates of the sentence ranker (MMR). However, it does not discard low-ranked ones, as the ranker is imperfect, and those sentences ranked low may also contribute to a high-quality summary. Soft attention restrains the search space while allowing more exploration of the limited labeled data, leading to better representation learning. Specifically, \ours infuses the entire prediction of MMR into 
its neural module by attending (restraining) to important sentences and downplaying the rest instead of completely discarding them.
(2) \ours resolves the \textbf{high redundancy} of MDS in a unified way:
the explicit redundancy measure in MMR is incorporated into the neural representation of the current state, and the two modules are coordinated by RL reward optimization, which encourages non-redundant summaries.

We conduct extensive experiments and ablation studies to examine the effectiveness of \ours.
Experimental results show that \ours achieves state-of-the-art performance on the DUC 2004~\cite{paul2004introduction} and TAC 2011~\cite{owczarzak2011overview} datasets (Sec.~\ref{sec:exp_sota}).
A comparison between various combination mechanisms demonstrates the benefits of soft attention in the large search space of MDS (Sec.~\ref{sec:exp_combine}).
In addition, ablation and manual studies confirm that \ours is superior to applying either RL or MMR to MDS alone, and MMR guidance is effective for redundancy avoidance (Sec.~\ref{sec:exp_mmr_guidance}).

\start{Contributions}
(1) We present an RL-based MDS framework that combines the advances of classical MDS and neural SDS methods via end-to-end learning. 
(2) We show that our proposed soft attention is better than the hard cutoff of previous methods for learning adequate neural representations. Also, infusing the neural representation of the current summary with explicit MMR measures significantly reduces summary redundancy.
(3) We demonstrate that \ours achieves new state-of-the-art results on benchmark MDS datasets.

\section{Problem Formulation}
\begin{figure*}[ht]
    \centering
    \includegraphics[width=0.78\linewidth]{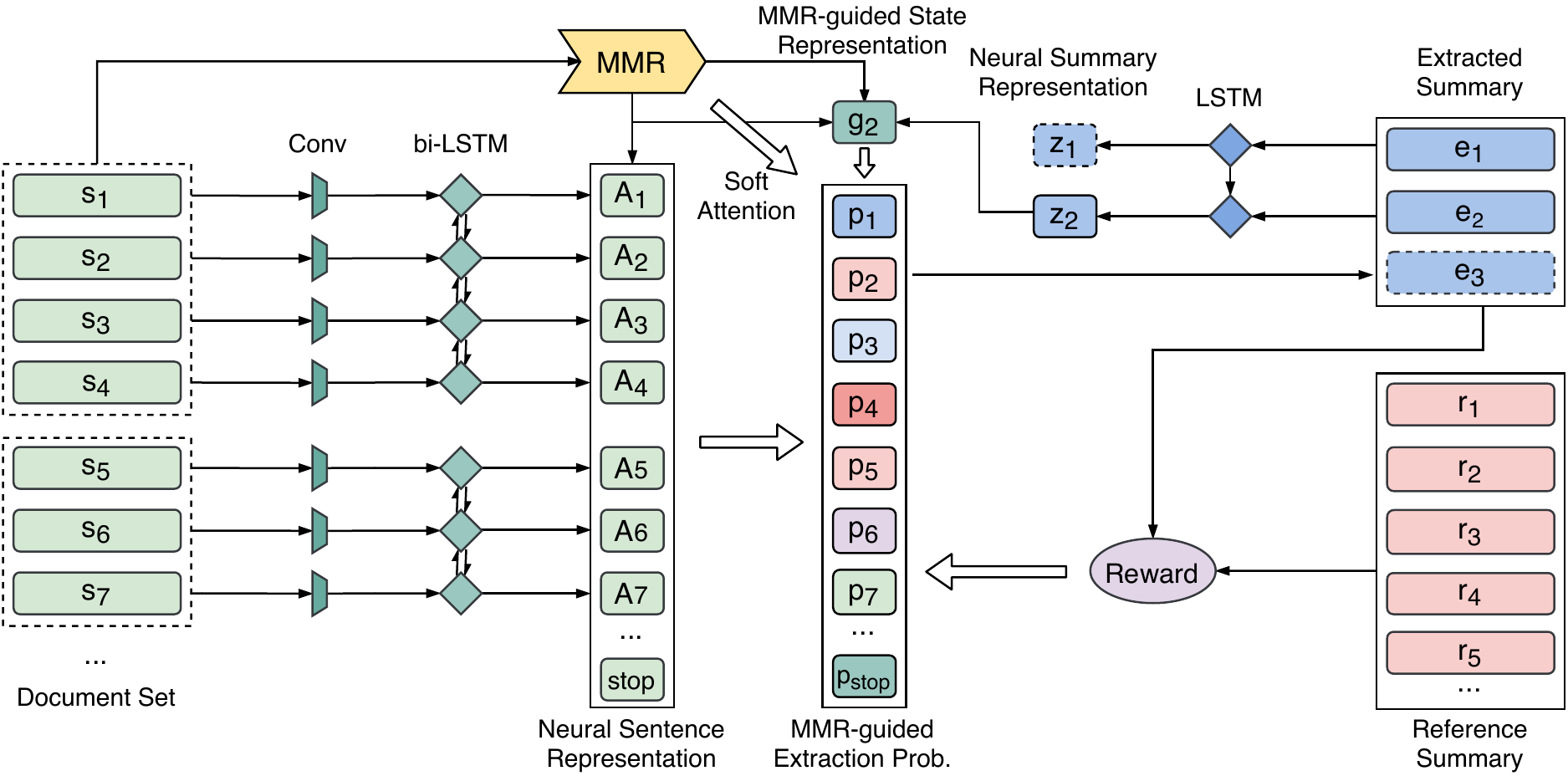}
    \upv
    \caption{\textbf{An overview of the proposed MDS framework \ours}. Neural sentence representation $\mA_j$ is obtained through sentence-level convolutional encoder and document-level bi-LSTM encoder. MMR guidance is incorporated into neural sentence representation $\mA_j$ and state representation $\vg_t$ through soft attention and end-to-end learned through reward optimization. }
    \label{fig:framework}
    \downv
\end{figure*}

We define $\gD = \{D_1, D_2, ..., D_N\}$ as a set of documents on the same topic.
Each document set $\gD$ is paired with a set of (human-written) reference summaries $\gR$.
For the convenience of notation, we denote the j-th sentence in $\gD$ as $s_j$ when concatenating the documents in $\gD$.
We focus on extractive summarization where a subset of sentences in $\gD$ are extracted as the system summary $\mE$.
A desired system summary $\mE$ covers salient and non-redundant information in $\gD$.
$\mE$ is compared with the reference summaries $\gR$ for evaluation.

\section{The \ours Framework}

\start{Overview}
At a high level, \ours infuses MMR guidance into end-to-end training of the neural summarization model.
\ours uses hierarchical encoding to efficiently encode the sentences in multiple documents and obtains the neural sentence representation $\mA_j$.
\ours models salience by combining MMR and sentence representation $\mA_j$, and measures redundancy by infusing MMR with neural summary representation $\vz_t$, which together form the state representation $\vg_t$.
At each time step, one sentence is extracted based on the MMR-guided sentence representation and state representation, and compared with the reference, the result (reward) of which is then back-propagated for the learning of both neural representation and MMR guidance.
An illustrative figure of \ours is shown in Fig.~\ref{fig:framework}.

In the following, we first describe MMR and the neural sentence representation. We then introduce the neural sentence extraction module and how we incorporate MMR guidance into it for better MDS performance. Finally, we illustrate how neural representation and MMR guidance are jointly learned via end-to-end reinforcement learning.

\subsection{Maximal Marginal Relevance}
\label{sec:mmr}
Maximal Marginal Relevance (MMR)~\cite{carbonell1998use} is a general summarization framework  that balances summary salience and redundancy.
Formally, MMR defines the score of a sentence $s_j$ at time $t$ as $m^t_j = \lambda \rmS(s_j, \gD ) - (1-\lambda) \max_{e \in \mE_t} \rmR(s_j, e)$,
where $\lambda \in [0, 1]$ is the weight balancing salience and redundancy.
$\rmS(s_j, \gD)$ measures how salient a sentence $s_j$ is, estimated by the similarity between $s_j$ and $\gD$. $\rmS(s_j, \gD)$ does not change during the extraction process.
$\mE_t$ consists of sentences that are already extracted before time $t$. $\max_{e \in \mE_t} \rmR(s_j, e)$ measures the redundancy between $s_j$ and each extracted sentence $e$ and finds the most redundant pair. $\max_{e \in \mE_t} \rmR(s_j, e)$ is updated as the size of $\mE_t$ increases.
Intuitively, if $s_j$ is similar to any sentence $e \in \mE_t$, it would be deemed redundant and less favored by MMR.
There are various options regarding the choices of $\rmS(s_j, \gD)$ and $\rmR(s_j, e)$, which we compare in Sec.~\ref{sec:diff_sim}.

We denote the (index of the) sentence extracted at time $t$ as $e_t$.
MMR greedily extracts one sentence at a time according to the MMR score: $e_t = \argmax_{s_j \in \gD \setminus \mE_t} m^t_j$.
Heuristic and deterministic algorithms like MMR are rather efficient and work reasonably well in some cases. However, they lack holistic modeling of summary quality and the capability of end-to-end representation learning.

\subsection{Neural Sentence Representation}
\label{sec:rep}
To embody end-to-end representation learning, we leverage the advances in SDS neural sequence learning methods.
Unlike prior studies on adapting SDS to MDS \cite{lebanoff-etal-2018-adapting}, which concatenates all the documents chronologically and encodes them sequentially, we %
adapt hierarchical encoding for better efficiency and scalability.
Specifically, we first encode each sentence $s_j$ via a CNN~\cite{kim-2014-convolutional} to obtain its sentence representation.
We then separately feed the sentence representations in \textit{each document} $D_i$ to a bi-LSTM~\cite{huang2015bidirectional}. %
The bi-LSTM generates a contextualized representation for each sentence $s_j$, denoted by $\vh_j$.
We form an action matrix $\mA$ using $\vh_j$, where the j-th row $\mA_j$ corresponds to the $j$-th sentence ($s_j$) in $\gD$.
A pseudo sentence indicating the STOP action, whose representation is randomly initialized, is also included in $\mA$, and sentence extraction is finalized when the STOP action is taken \cite{mao-etal-2018-end,DBLP:conf/emnlp/MaoTHR19}. %

\subsection{Neural Sentence Extraction}
We briefly describe the SDS sentence extraction module~\cite{chen-bansal-2018-fast} that we base our work on, and elaborate in Sec~\ref{sec:rl-mmr} how we adapt it for better MDS performance with MMR guidance.

The probability of neural sentence extraction is measured through a two-hop attention mechanism.
Specifically, we first obtain the neural summary representation $\vz_t$ by feeding previously extracted sentences ($\mA_{e_{i}}$) to an LSTM encoder.
 A time-dependent state representation $\vg_t$ that considers both sentence representation $\mA_j$ and summary representation $\vz_t$ is obtained by the \textit{glimpse} operation~\cite{vinyals2015order}:

\small
\begin{align} \label{eq:glimpse}
\upve
  a^t_j &=
    \vv_1^\intercal \tanh(\mW_{1} \mA_j + \mW_{2} \vz_t), \\
  {\boldsymbol\alpha} ^t &= \text{softmax}(\va^t), \\
  \vg_t &= \sum_j {\boldsymbol\alpha}^t_j \mW_{1} \mA_j,
 \downve
\end{align}
\normalsize
where $\mW_1$, $\mW_2$, $\vv_1$ are model parameters.
$\va^t$ represents the vector composed of $a^t_j$.
With $\vz_t$, the attention weights ${\boldsymbol\alpha}^t_j$ are aware of previous extraction. %
Finally, the sentence representation $\mA_j$ is attended again to estimate the extraction probability.
\begin{equation} 
\label{eq:ptrnet}
\small
\upve
p^t_j = \begin{cases}
  \ \vv_2^{\intercal} \tanh(
  \mW_{3} \mA_j + \mW_{4} \vg_t) & 
        \text{if } s_j \ne e_i, \forall i < t \\
  \ -\infty & \text{otherwise},
\end{cases}
\downve
\end{equation}
where $\mW_3$, $\mW_4$, $\vv_2$ are model parameters and previously extracted sentences $\{ e_i\}$ are excluded.

The summary redundancy here is handled implicitly by $\vg_t$. 
Supposedly, a redundant sentence $s_j$ would receive a low attention weight $a^t_j$ after comparing $\mA_j$ and $\vz_t$ in Eq.~\ref{eq:glimpse}.
However, we find such latent modeling insufficient for MDS due to its much higher degree of redundancy.
For example, when news reports start with semantically similar sentences, using latent redundancy avoidance alone leads to repeated summaries (App \ref{sec:details} Table~\ref{tab:redundancy}). Such observations motivate us to incorporate MMR, which models redundancy explicitly, to guide the learning of sentence extraction for MDS.

\subsection{MMR-guided Sentence Extraction}
\label{sec:rl-mmr}
In this section, we describe several strategies of incorporating MMR into sentence extraction, which keeps the neural representation for expressiveness while restraining the search space to fewer promising candidates for more adequate representation learning under limited training data.

\start{Hard Cutoff}
One straightforward way of incorporating MMR guidance is to only allow extraction from the top-ranked sentences of MMR.
We denote the sentence list ranked by MMR scores $m^t_j$ as $\mM^t$.
Given $p^t_j$ -- the neural probability of sentence extraction before softmax, we set the probability of the sentences after the first $K$ sentences in $\mM^t$ to $-\infty$.
In this way, the low-ranked sentences in MMR are never selected and thus never included in the extracted summary.
We denote this variant as \oursHard.

There are two limitations of conducting hard cutoff in the hope of adequate representation learning:
\textbf{(L1)} Hard cutoff ignores the values of MMR scores and simply uses them to make binary decisions.
\textbf{(L2)} While hard cutoff reduces the search space, the decision of the RL agent is limited as it cannot extract low-ranked sentences and thus lacks exploration of the (already limited) training data.
To tackle L1, a simple fix is to \textit{combine} the MMR score $m^t_j$ with the extraction probability measured by the neural sentence representation.
\begin{equation} \label{eq:topk-mlp}
\small
\upve
p^t_j = \begin{cases}
  \ \beta \vv_2^\intercal \tanh(
  \mW_{3} \mA_j + \mW_{4} \vg_t) + (1 - \beta) \text{FF}(m^t_j) \\  
        \quad\quad\quad\quad \text{if } s_j \ne e_i, \forall i < t\ \text{and}\ s_j \in \mM^t_{1:K}  \\
  \ -\infty \quad\quad\ \text{otherwise},
\end{cases} \\
\downve
\end{equation}
where $\beta \in [0, 1]$ is a constant. $\text{FF} (\cdot)$ is a two-layer feed-forward network that enables more flexibility than using raw MMR scores, compensating for the magnitude difference between the two terms. We denote this variant as \oursHardComb.

\start{Soft Attention}
To deal with L2, we explore soft variants that do not completely discard the low-ranked sentences but encourage the extraction of top-ranked sentences.
The first variant, \oursSoftComb, removes the constraint of $s_j \in \mM^t_{1:K}$ in Eq.~\ref{eq:topk-mlp}. This variant solves L2 but may re-expose the RL agent to L1 since its MMR module and neural module are loosely coupled and there is a learnable layer in their combination.

Therefore, we design a second variant, \oursSoftAttn, which addresses both L1 and L2 by tightly incorporating MMR into neural representation learning via soft attention.
Specifically, the MMR scores are first transformed and normalized: ${\boldsymbol\mu}^t = \softmax (\text{FF}(\vm^t))$, and then used to attend neural sentence representation $\mA_j$ before the two-hop attention: $\mA'_j = {\boldsymbol\mu}^t_j \mA_j$. 
The state representation $\vg_t$, which captures summary redundancy, is also impacted by MMR through the attention between summary representation $\vz_t$ and MMR-guided sentence representation $\mA'_j$ in Eq.~\ref{eq:glimpse}.
L1 is addressed as ${\boldsymbol\mu}^t$ represents the extraction probability estimated by MMR.
L2 is resolved since the top-ranked sentences in MMR receive high attention, which empirically is enough to restrain the decision of the RL agent, while the low-ranked sentences are downplayed but not discarded, allowing more exploration of the search space.

\subsection{MDS with Reinforcement Learning}
\label{sec:rl}
The guidance of MMR is incorporated into neural representation learning  through end-to-end RL training.
Specifically, we formulate extractive MDS as a Markov Decision Process, where the state is defined by $(\gD \setminus \mE_t, \vg_t)$.
At each time step, one action is sampled from $\mA$ given $p^t_j$, and its reward is measured by  
comparing the extracted sentence $e_t$ with the reference $\gR$ via ROUGE \cite{lin-2004-rouge}, \ie, $r_t = \text{ROUGE-L}_{\text{F1}} (e_t, \gR)$.
At the final step $T$ when the STOP action is taken, 
an overall estimation of the summary quality is measured by $r_T = \text{ROUGE-1}_{\text{F1}} (\mE, \gR)$.
Reward optimization encourages salient and non-redundant summaries -- intermediate rewards focus on the sentence salience of the current extracted sentence and the final reward captures the salience and redundancy of the entire summary.

Similar to prior RL-based models on SDS~\cite{paulus2017deep,chen-bansal-2018-fast,narayan-etal-2018-ranking}, we use policy gradient~\cite{williams1992simple} as the learning algorithm for model parameter updates.
In addition, we adopt Advantage Actor-Critic (A2C) optimization -- a critic network is added to enhance the stability of vanilla policy gradient. 
The critic network has a similar architecture to the one described in Sec.~\ref{sec:rep} and uses the sentence representation $\mA$ to generate an estimation of the discounted reward, which is then used as the baseline subtracted from the actual discounted reward before policy gradient updates.

\section{Experiments}
We conduct extensive experiments to examine \ours with several key questions:
\textbf{(Q1)} How does \ours perform compared to state-of-the-art methods?
\textbf{(Q2)} What are the advantages of soft attention over hard cutoff in learning adequate neural representations under the large search space?
\textbf{(Q3)} How crucial is the guidance of MMR for adapting SDS to MDS in the face of high redundancy?

\subsection{Experimental Setup}

\start{Datasets}
We take the MDS datasets from DUC and TAC competitions which are widely used in prior studies~\cite{kulesza2012determinantal,lebanoff-etal-2018-adapting}.
Following convention~\cite{wang-etal-2017-affinity,cao2017improving,cho-etal-2019-improving}, DUC 2004 (trained on DUC 2003) and TAC 2011 (trained on TAC 2008-2010) are used as the test sets.
We use DUC 2004 as the validation set when evaluated on TAC 2011 and vice versa.
More details of the dataset statistics are in App.~\ref{app:data}.

\start{Evaluation Metrics}
In line with recent work~\cite{li2017salience,lebanoff-etal-2018-adapting,zhang-etal-2018-adapting,cho-etal-2019-improving}, we measure ROUGE-1/2/SU4 F1 scores \cite{lin-2004-rouge}.
The evaluation parameters are set according to~\citet{hong-etal-2014-repository} with stemming and stopwords not removed.
The output length is limited to 100 words. These setups are the same for all compared methods.\footnote{Parameters of ROUGE: -2 4 -U -r 1000 -n 2 -l 100 -m.}

\start{Compared Methods}
We compare \ours with both classical and neural MDS methods.
Note that \textit{some previous methods are incomparable} due to differences such as length limit (100 words or 665 bytes) and evaluation metric (ROUGE F1 or recall).
Details of each method and differences in evaluation can be found in App.~\ref{app:baseline}.

For extractive methods, we compare with SumBasic~\cite{vanderwende2007beyond}, {KLSumm}~\cite{haghighi-vanderwende-2009-exploring}, {LexRank}~\cite{erkan2004lexrank}, {Centroid}~\cite{hong-etal-2014-repository}, {ICSISumm}~\cite{gillick-favre-2009-scalable},
{rnn-ext + RL}~\cite{chen-bansal-2018-fast}, {DPP}~\cite{kulesza2012determinantal}, and DPP-Caps-Comb~\cite{cho-etal-2019-improving}.
For abstractive methods, we compare with {Opinosis}~\cite{ganesan-etal-2010-opinosis}, {Extract+Rewrite}~\cite{song-etal-2018-structure}, {PG}~\cite{see-etal-2017-get}, and {PG-MMR}~\cite{lebanoff-etal-2018-adapting}.

We use \oursSoftAttn as our default model unless otherwise mentioned.
Implementation details can be found in App.~\ref{sec:implementation}.
We also report Oracle, an approximate upper bound that greedily extracts sentences to maximize ROUGE-1 F1 given the reference summaries~\cite{nallapati2017summarunner}.

\subsection{Experimental Results}

\subsubsection{Comparison with the State-of-the-art}
\label{sec:exp_sota}

To answer Q1, we compare \ours with state-of-the-art summarization methods and list the comparison results in Tables~\ref{tab:results_duc04} and \ref{tab:results_tac11}.

On DUC 2004, we observe that rnn-ext + RL, which we base our framework on, fails to achieve satisfactory performance even after fine-tuning.
The large performance gains of \ours over rnn-ext + RL demonstrates the benefits of guiding SDS models with MMR when adapting them to MDS. 
A similar conclusion is reached when comparing PG and PG-MMR. However, the hard cutoff in PG-MMR and the lack of in-domain fine-tuning lead to its inferior performance.
DPP and DPP-Caps-Comb obtain decent performance but could not outperform \ours due to the lack of end-to-end representation learning.
Lastly, \ours achieves new state-of-the-art results, approaching the performance of Oracle, which has access to the reference summaries, especially on ROUGE-2.
We observe similar trends on TAC 2011 in which \ours again achieves state-of-the-art performance. The improvement over compared methods is especially significant on ROUGE-1 and ROUGE-SU4.

\begin{table}[t]
\centering

\resizebox{.84\columnwidth}{!}{
\begin{tabular}{l rrr}
\toprule
\multirow{2}{*}{\textbf{Method}}& \multicolumn{3}{c}{\textbf{DUC 2004}}\\
 & \textbf{R-1} & \textbf{R-2} & \textbf{R-SU4} \\
\midrule
Opinosis & 27.07 & 5.03 & 8.63 \\
Extract+Rewrite & 28.90 & 5.33 & 8.76 \\
SumBasic & 29.48 & 4.25 & 8.64\\
KLSumm & 31.04 & 6.03 & 10.23 \\
LexRank & 34.44 & 7.11 & 11.19 \\
Centroid & 35.49 & 7.80 & 12.02 \\
ICSISumm & 37.31 & 9.36 & 13.12 \\
PG & 31.43 & 6.03 & 10.01\\
PG-MMR & 36.88 & 8.73 & 12.64\\
{rnn-ext + RL (pre-train)} & 32.76 & 6.09 & 10.36\\
{rnn-ext + RL (fine-tune)} & 35.93 & 8.60 & 12.53\\
DPP & 38.10 & 9.14 & 13.40 \\
DPP-Caps-Comb\textdagger & 37.97 & 9.68 & 13.53\\
\ours (ours) & \textbf{38.56} & \textbf{10.02} & \textbf{13.80}\\
\midrule

Oracle & 39.67 & 10.07 & 14.31 \\
\bottomrule
\end{tabular}
}

\upv
\caption{\textbf{ROUGE F1 of compared methods on DUC 2004}. \textdagger We re-evaluate DPP-Caps-Comb~\cite{cho-etal-2019-improving} using author-released output as we found its  results did not follow the 100-word length limit.
}
\label{tab:results_duc04}
\downv
\end{table}

\begin{table}[t]
\centering

\resizebox{.84\columnwidth}{!}{%
\begin{tabular}{lrrr}
\toprule
\multirow{2}{*}{\textbf{Method}}& \multicolumn{3}{c}{\textbf{TAC 2011}}\\
 & \textbf{R-1} & \textbf{R-2} & \textbf{R-SU4} \\
\midrule
Opinosis & 25.15 & 5.12 & 8.12\\
Extract+Rewrite & 29.07 & 6.11 & 9.20\\
SumBasic & 31.58 & 6.06 & 10.06\\
KLSumm & 31.23 & 7.07 & 10.56 \\
LexRank & 33.10 & 7.50 & 11.13 \\
PG & 31.44 & 6.40 & 10.20\\
PG-MMR & 37.17 & 10.92 & 14.04\\
{rnn-ext + RL (pre-train)} & 33.45 & 7.37 & 11.28\\
{rnn-ext + RL (fine-tune)} & 37.13 & 10.72 & 14.16\\
DPP & 36.95 & 9.83 & 13.57 \\
DPP-Caps-Comb\textdagger & 37.51 & 11.04 & 14.16 \\
\ours (ours) & \textbf{39.65} & \textbf{11.44} & \textbf{15.02}\\
\midrule

Oracle &  42.44&  13.85 & 16.90 \\
\bottomrule
\end{tabular}
}

\upv
\caption{\textbf{Results of automatic evaluation (ROUGE F1) on TAC 2011}. \textdagger The output of DPP-Caps-Comb is again re-evaluated by limiting to 100 words.
}
\label{tab:results_tac11}
\downv
\end{table}

\subsubsection{Analysis of \ours Combination}\label{sec:exp_combine}
We answer Q2 by comparing the performance of various combination mechanisms for \ours.

\start{Performance Comparison}
As shown in Table~\ref{tab:ablation_duc04},
\oursHardComb performs better than \oursHard, showing the effectiveness of using MMR scores instead of degenerating them into binary values.
We test \oursSoftComb with different $\beta$ but it generally performs much worse than other variants, which implies that naively incorporating MMR into representation learning through weighted average may loosen the guidance of MMR, losing the benefits of both modules.
Infusing MMR via soft attention of the action space performs the best, demonstrating the effectiveness of MMR guidance in \oursSoftAttn for sentence representation learning.

\begin{table}[ht]
\centering

\scalebox{.76}{
\begin{tabular}{l rrr}
\toprule
\multirow{2}{*}{\textbf{Combination}}& \multicolumn{3}{c}{\textbf{DUC 2004}}\\
& \textbf{R-1} & \textbf{R-2} & \textbf{R-SU4} \\
\midrule
\Hard & 38.19 & 9.26 & 13.43 \\ %
\HardComb & 38.45 & 9.35 & 13.64 \\ 
\SoftComb & 37.70 & 8.90 & 12.98\\ 
\SoftAttn & \textbf{38.56} & \textbf{10.02} & \textbf{13.80}\\
\bottomrule
\end{tabular}
}

\upv
\caption{\textbf{Comparison of \ours variants} with different combination mechanisms.}
\label{tab:ablation_duc04}
\downv
\end{table}

\start{Hard Cutoff vs. Soft Attention}
We further compare the extracted summaries of MMR, \oursHard, and \oursSoftAttn to verify the assumption that there are high-quality sentences not ranked highly by MMR and thus neglected by the hard cutoff.
In our analysis, we find that when performing soft attention, 32\% (12\%) of extracted summaries contain low-ranked sentences that are not from $\mM^1_{1:K}$ when $K = 1$ ($K = 7$).
We then evaluate those samples with low-ranked sentences extracted and conduct a pairwise comparison.
On average, we observe a gain of 18.9\% ROUGE-2 F1 of \oursSoftAttn over MMR, and 2.71\% over \oursHard, which demonstrates the benefits of soft attention.

\start{Degree of \ours Combination}
To study the effect of \ours combination in different degrees, we vary the cutoff $K$ in \oursHard and analyze performance changes.
As listed in Table~\ref{tab:diff_k}, a small $K(=1)$ imposes tight constraints and practically degrades \ours to vanilla MMR.
A large $K(=50)$ might be too loose to limit the search space effectively, resulting in worse performance than a $K(=7, 10)$ within the proper range.
When $K$ is increased to 100, the impact of MMR further decreases but still positively influences model performance compared to the vanilla RL ($K = \infty$), especially on ROUGE-1.

\begin{table}[ht]
\centering

\resizebox{.9\columnwidth}{!}{%
\begin{tabular}{l rrr rrr}
\toprule
\multirow{2}{*}{\textbf{K}} & \multicolumn{3}{c}{\textbf{DUC 2004}} & \multicolumn{3}{c}{\textbf{TAC 2011}}\\
 & \textbf{R-1} & \textbf{R-2} & \textbf{R-SU4} & \textbf{R-1} & \textbf{R-2} & \textbf{R-SU4}\\
\midrule
1 & 37.91 & 8.83 & 13.10 & 38.54 & 10.83 & 14.43\\
7 & 38.19 & \textbf{9.26} & 13.43 & \textbf{39.22} & \textbf{11.10} & \textbf{14.78}\\
10 & \textbf{38.22} & 9.24 & \textbf{13.49} & 39.13 & 11.07  & 14.63\\ %
50 & 38.12 & 9.23  & 13.42  & 38.60& 11.05 & 14.55 \\ %
100 & 36.92 & 8.98 & 12.98 & 37.94 & 10.92  & 14.20\\  %
$\infty$ & 35.93 & 8.60 & 12.53 & 37.13& 10.72& 14.16\\
\bottomrule

\end{tabular}
}

\upv
\caption{\textbf{Performance changes of \oursHard} when different cutoffs ($K$) are used.}
\label{tab:diff_k}
\downv
\end{table}

\subsubsection{Effectiveness of MMR Guidance}\label{sec:exp_mmr_guidance}
To answer Q3, we compare \ours with vanilla RL without MMR guidance in terms of both training and test performance.
We also inspect details such as runtime and quality of their extracted summaries (provided in App.\ref{sec:details}).

\begin{figure}[ht]
    \centering
    \includegraphics[width=0.75\linewidth]{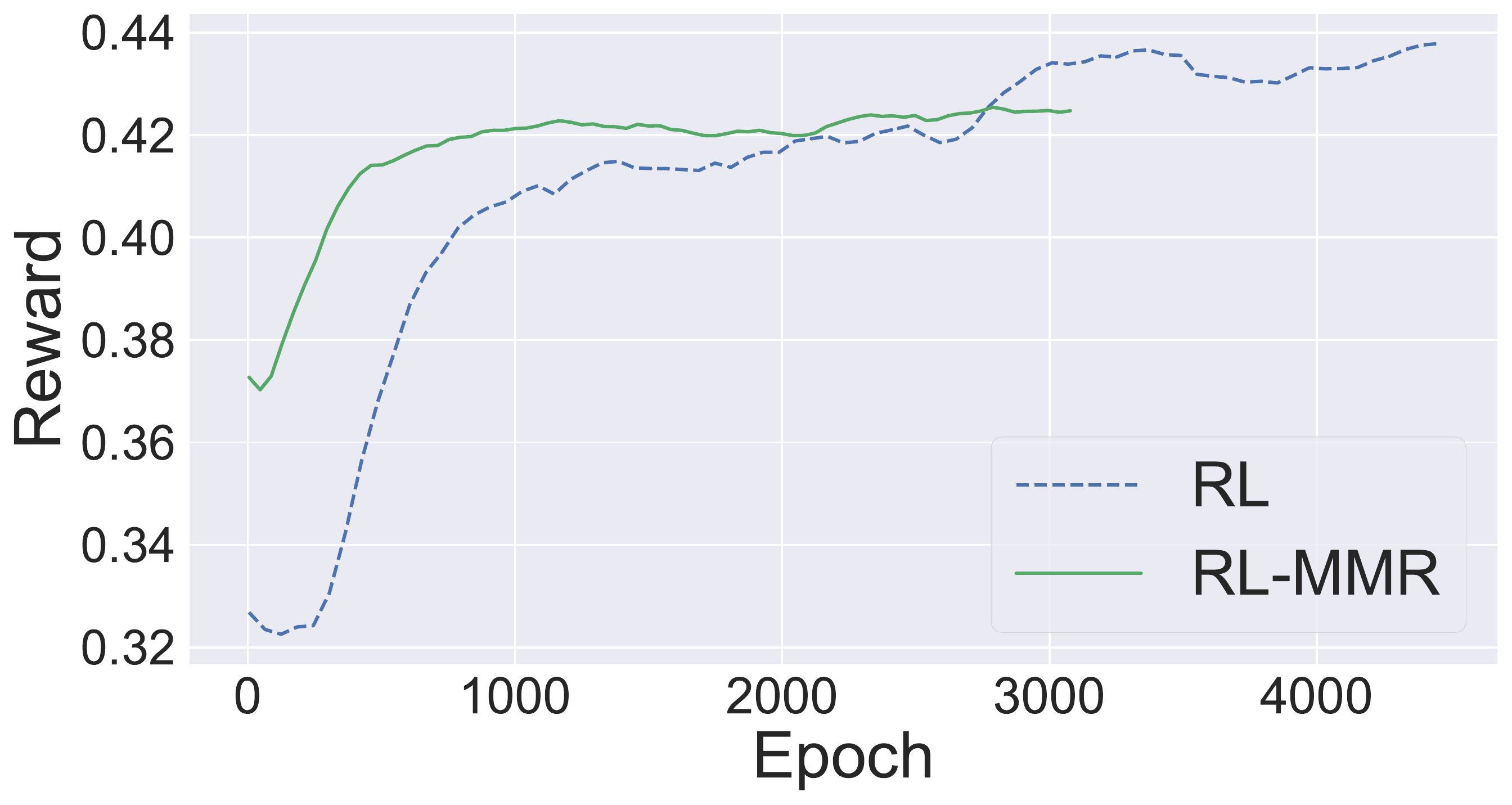}
    \upv
    \caption{\textbf{The learning curves of RL and \ours on DUC 2004}. RL training is more stable and converges faster when equipped with MMR.}
    \label{fig:learning_curves}
    \downv
\end{figure}

\start{Training Performance}
To examine whether MMR guidance helps with the learning efficiency of MDS, we plot the learning curves of vanilla RL and \oursHard in Fig.~\ref{fig:learning_curves}.
\ours receives a significantly better initial reward on the training set because MMR provides prior knowledge to extract high-quality sentences.
In addition, \ours has lower variance and achieves faster convergence than RL due to MMR guidance. 
Note that the final reward of vanilla RL on the plateau is higher than \ours, which is somewhat expected since RL can achieve better fitting on the training set when it has less guidance (constraint).

\start{Test Performance}
We compare the test performance of vanilla RL and \ours in Table~\ref{tab:rl_vs_rl-mmr}.
Despite the fact that vanilla RL obtains better training performance, its test performance is significantly worse than \ours.
Such a contradiction indicates that vanilla RL overfits on the training set and does not generalize well, again demonstrating the benefits of MMR guidance.
We also find that, perhaps surprisingly, MMR outperforms vanilla RL even when RL is fine-tuned using in-domain training data. 
We thus believe that MMR and its methodology are underestimated by prior studies and should be explored further.
Finally, \ours achieves significantly better results than either RL or MMR alone, demonstrating the superiority of combining RL with MMR for MDS.

\begin{table}[ht]
\centering

\resizebox{\columnwidth}{!}{%
\begin{tabular}{l rrr rrr}
\toprule
\multirow{2}{*}{\textbf{Method}} & \multicolumn{3}{c}{\textbf{DUC 2004}} & \multicolumn{3}{c}{\textbf{TAC 2011}}\\
 & \textbf{R-1} & \textbf{R-2} & \textbf{R-SU4} & \textbf{R-1} & \textbf{R-2} & \textbf{R-SU4}\\
\midrule
RL (pre-train) & 32.76 & 6.09 & 10.36 & 33.45 & 7.37 & 11.28\\
RL (fine-tune) & 35.93 & 8.60 & 12.53 & 37.13 & 10.72 & 14.16\\
MMR & 37.90 & 8.83 & 13.10 & 38.53 & 10.83 & 14.44\\
\ours  & \textbf{38.56} & \textbf{10.02} & \textbf{13.80} & \textbf{39.65} & \textbf{11.44} & \textbf{15.02}\\ 
\bottomrule

\end{tabular}
}

\upv
\caption{\textbf{Comparison of MMR, RL, and \ours} further shows the effectiveness of MMR guidance.}
\label{tab:rl_vs_rl-mmr}
\downv
\end{table}

\subsubsection{Ablation of MMR}\label{sec:diff_sim}
In this section, we conduct more ablation of MMR given its decent performance.
We study the balance between salience and redundancy, and the performance of different similarity measures.
Specifically, we use TF-IDF and BERT \cite{devlin2018bert} as the sentence (document) representation and measure cosine similarity in $\rmS(s_j, \gD)$ and $\rmR(s_j, e)$. 
We also explore whether a semantic textual similarity model, SNN~\cite{siamese-github}, is more effective in measuring redundancy $\rmR(s_j, e)$ than TF-IDF.
The TF-IDF features are estimated on the MDS datasets while the neural models are pre-trained on their corresponding tasks.

\begin{table*}[t]
    \footnotesize
    \centering
        \scalebox{.97}{
        \scalebox{.933}{
        \begin{minipage}[b]{0.49\hsize}%
        \begin{tabular}{|p{.97\columnwidth}}
            \toprule
             \textbf{Reference Summary}: PKK leader Ocalan was {arrested on arrival at the Rome airport}.
He asked for asylum.
\colorB{Turkey pressured Italy to extradite Ocalan}, whom they consider a terrorist.
\colorB{Kurds in Europe flocked to Rome to show their support}.
\colorB{About 1,500 staged a hunger strike outside the hospital where he was held}.
Italy began a border crackdown to stop Kurds flocking to Rome.
\colorB{Greek media and officials oppose extradition}.
\colorB{Romanian Kurds staged a 1-day business shutdown to protest his arrest}.
In a Turkish prison, an Italian prisoner was taken hostage.
The Turkish president needed extra security for a trip to Austria.
This is Italy's Prime Minister D'Alema's first foreign policy test. \\
             \midrule
             \textbf{DPP-Caps-Comb}: 1. \colorB{Turkey has asked for his extradition and Ocalan has asked for political asylum.}  \\
2. Turkey \colorR{stepped up the pressure on Italy for the extradition} of captured Kurdish rebel leader Abdullah Ocalan, warning Sunday that granting him asylum would amount to opening doors to terrorism.\\
3. \colorG{If Italy sends Ocalan back to Turkey, he'll be tortured for certain}, said Dino Frisullo, an Italian supporter among the singing, chanting Kurds outside the military hospital.\\
4. \colorB{Thousands of Kurds living in Romania closed down restaurants, shops and companies to protest the arrest} of leader Abdullah Ocalan by Italian authorities, a newspaper reported Tuesday.\\
5. \colorR{Turkey wants Italy 
to extradite the rebel, Abdullah Ocalan}, leader of the Kurdistan Workers' 
Party, which is seeking Kurdish autonomy in southeastern Turkey.\\
            \bottomrule
        \end{tabular}
        \end{minipage}
}
        
\begin{minipage}[b]{0.52\hsize}%
        \begin{tabular}{|p{.97\columnwidth}|}
            \toprule
 \textbf{MMR}: 1. \colorB{Turkey wants Italy 
to extradite the rebel, Abdullah Ocalan}, leader of the Kurdistan Workers' 
Party, which is seeking Kurdish autonomy in southeastern Turkey. \\
2. Earlier Monday, while members of D'Alema's government 
met with Turkish officials who were in Rome for a European ministerial 
meeting, \colorB{thousands of Kurds flooded into Rome} to hold a demonstration 
and \colorB{hunger strike in support of Ocalan}.
\\
3. The \colorB{extra effort was prompted} by the arrest last 
week in Rome of Abdullah Ocalan, the chief of the Turkish Workers 
Party PKK, Zehetmayr said.\\
4. \colorG{If Italy sends the Kurd leader back to Turkey, he'll be tortured 
for certain}, said Dino Frisullo, an Italian supporter among the 
singing, chanting Kurds outside the military hospital.\\             
             \midrule
\textbf{\ours}: 1. \colorB{Turkey wants Italy 
to extradite the rebel, Abdullah Ocalan}, leader of the Kurdistan Workers' 
Party, which is seeking Kurdish autonomy in southeastern Turkey. \\

2. \colorB{In Rome, 1,500 Kurds massed} for a second day of demonstrations \colorB{outside the 
military hospital where Ocalan is believed to be held.}\\

3. \colorB{Thousands of Kurds living in Romania closed down restaurants, shops 
and companies to protest the arrest} of leader Abdullah Ocalan by Italian 
authorities, a newspaper reported Tuesday.\\

4. \colorB{Greek media and officials leveled strong opposition Sunday to the 
possible extradition} of Abdullah Ocalan, the arrested Kurdish guerrilla 
leader, to Greece's traditional rival Turkey. \\

            \bottomrule
        \end{tabular}
        \end{minipage}

        }
       
            \upv
            \caption{\textbf{System summaries of different methods}. Text spans matched (unmatched) with the reference summary are in \colorB{blue} (\colorG{green}). Redundant spans are in \colorR{red}. Spans of the reference covered by \ours are also in \colorB{blue}.}
            \label{tab:casestudy}
             \downv
\end{table*}
\begin{figure}[ht]
    \includegraphics[width=.99\linewidth]{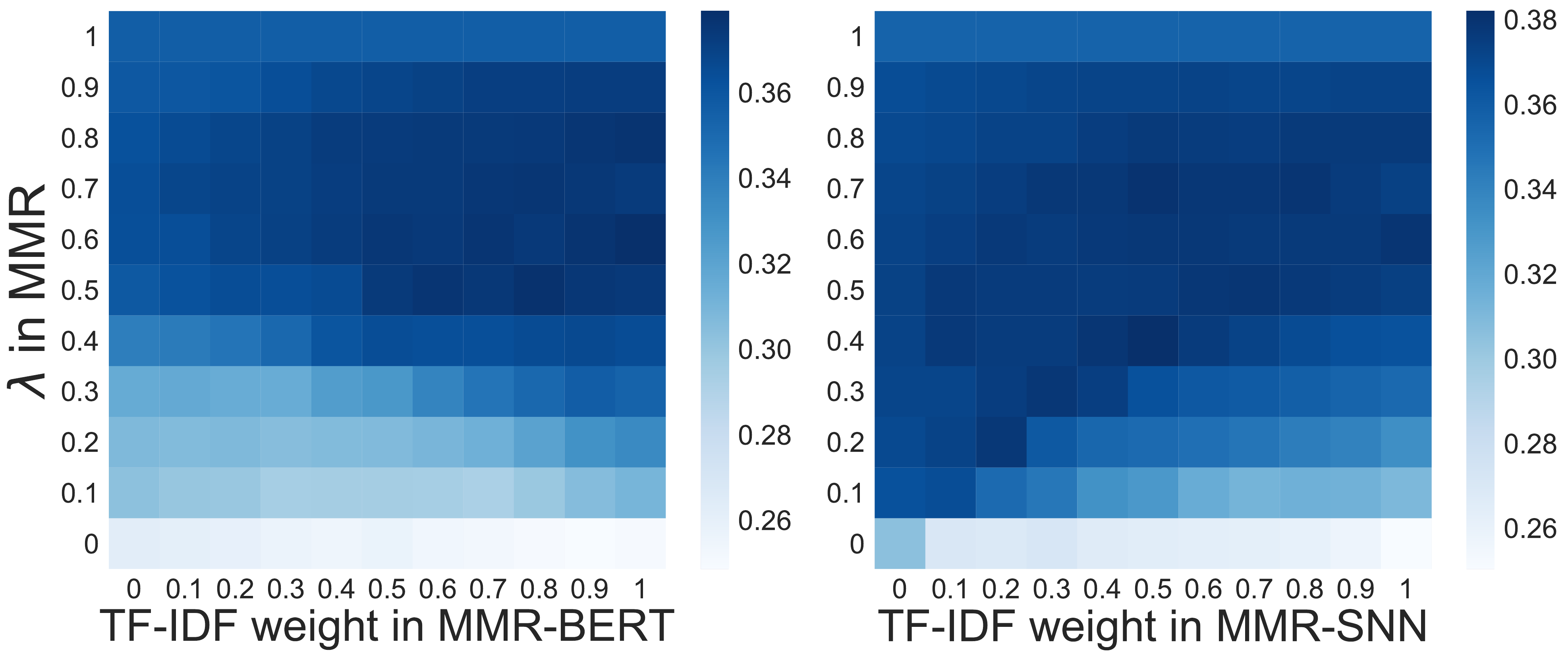}
    \upv
    \caption{\textbf{Performance comparison under different salience-redundancy balances (y-axis), and various weighted combinations} of TF-IDF and BERT \figleft\ or SNN \figright\ for similarity measure (x-axis). DUC 2004, ROUGE-1 F1, and $\rmR(s_j, e)$ are used for illustration. The results of other setups are similar.}
    \label{fig:grid_mmr}
    \downv
\end{figure}

\start{Balance between Salience and Redundancy}
By examining the y-axis in Fig.~\ref{fig:grid_mmr}, we observe that considering both salience and redundancy (best $\lambda =$ 0.5 \textasciitilde \ 0.8) performs much better than only considering salience ($\lambda = 1$) regardless of the specific measures, further indicating the necessity of explicit redundancy avoidance in MDS.

\start{Comparison of Similarity Measures}
By varying the x values in Fig.~\ref{fig:grid_mmr}, TF-IDF and neural estimations are combined using different weights.
Although BERT and SNN (combined with TF-IDF) perform \textit{slightly} better at times, they often require careful hyper-parameter tuning (both x and y).
Hence, We use TF-IDF as the representation in MMR throughout our experiments.

\subsubsection{Output Analysis}
We analyze the outputs of the best-performing methods in Table~\ref{tab:casestudy}.
DPP-Caps-Comb still seems to struggle with redundancy as it extracts three sentences with similar semantics (``Turkey wants Italy to extradite Ocalan'').
MMR and DPP-Caps-Comb both extract one sentence regarding a hypothesis that ``Ocalan will be tortured'', which is not found in the reference.
\ours has a more salient and non-redundant summary, as it is end-to-end trained with advances in SDS for sentence representation learning while maintaining the benefits of classical MDS approaches.
In contrast, MMR alone only considers lexical similarity; The redundancy measure in DPP-Caps-Comb is pre-trained on one SDS dataset with weak supervision and fixed during the training of DPP.

\section{Related Work}

\start{Multi-document Summarization}
Classical MDS explore both extractive \cite{ erkan2004lexrank,  haghighi-vanderwende-2009-exploring} and abstractive methods~\cite{barzilay-etal-1999-information, ganesan-etal-2010-opinosis}.
Many neural MDS methods \cite{yasunaga-etal-2017-graph, zhang-etal-2018-adapting} are merely comparable or even worse than classical methods due to the challenges of large search space and limited training data.
Unlike DPP-Caps-Comb~\cite{cho-etal-2019-improving} that incorporates neural measures into classical MDS as features, \ours opts for the opposite by endowing SDS methods with the capability to conduct MDS, enabling the potential of further improvement with advances in SDS.

\start{Bridging SDS and MDS}
Initial trials adapting SDS models to MDS~\cite{lebanoff-etal-2018-adapting,zhang-etal-2018-adapting} directly reuse SDS models~\cite{see-etal-2017-get,tan-etal-2017-abstractive}.
To deal with the large search space, a sentence ranker is used in the adapted models for candidate pruning. 
Specifically, \citet{lebanoff-etal-2018-adapting} leverages  MMR~\cite{carbonell1998use} to rank sentences, allowing only the words in the top-ranked sentences to appear in the generated summary. Similarly, \citet{zhang-etal-2018-adapting} uses topic-sensitive PageRank~\cite{haveliwala2002topic} and computes attention only for the top-ranked sentences.
Unlike \ours, these adapted models use hard cutoff and (or) lack end-to-end training, failing to outperform state-of-the-art methods designed specifically for MDS \cite{gillick-favre-2009-scalable, kulesza2012determinantal}.

\section{Conclusion}
We present a reinforcement learning framework for MDS that unifies neural SDS advances and Maximal Marginal Relevance (MMR) through end-to-end learning.
The proposed framework leverages the benefits of both neural sequence learning and statistical measures, bridging the gap between SDS and MDS.
We conduct extensive experiments on benchmark MDS datasets and demonstrate the superior performance of the proposed framework, especially in handling the large search space and high redundancy of MDS.
In the future, we will investigate the feasibility of incorporating classical MDS guidance to abstractive models with large-scale pre-training \cite{gu2020generating} and more challenging settings where each document set may contain hundreds or even thousands of documents.

\section*{Acknowledgments}
We thank Yu Zhang and Yu Meng for their help at the early stage of the work.
Research was sponsored in part by US DARPA KAIROS Program No. FA8750-19-2-1004 and SocialSim Program No.  W911NF-17-C-0099, National Science Foundation IIS-19-56151, IIS-17-41317, IIS 17-04532, and IIS 16-18481, and DTRA HDTRA11810026. Any opinions, findings, and conclusions or recommendations expressed herein are those of the authors and should not be interpreted as necessarily representing the views, either expressed or implied, of DARPA or the U.S. Government. The U.S. Government is authorized to reproduce and distribute reprints for government purposes notwithstanding any copyright annotation hereon.

\bibliography{multi-summ}

\begin{thebibliography}{48}
\expandafter\ifx\csname natexlab\endcsname\relax\def\natexlab#1{#1}\fi

\bibitem[{Barzilay et~al.(1999)Barzilay, McKeown, and
  Elhadad}]{barzilay-etal-1999-information}
Regina Barzilay, Kathleen~R. McKeown, and Michael Elhadad. 1999.
\newblock \href {https://doi.org/10.3115/1034678.1034760} {Information fusion
  in the context of multi-document summarization}.
\newblock In \emph{27th Annual Meeting of the Association for Computational
  Linguistics, University of Maryland, College Park, Maryland, USA, 20-26 June
  1999}, pages 550--557. {ACL}.

\bibitem[{Cao et~al.(2017)Cao, Li, Li, and Wei}]{cao2017improving}
Ziqiang Cao, Wenjie Li, Sujian Li, and Furu Wei. 2017.
\newblock \href {http://aaai.org/ocs/index.php/AAAI/AAAI17/paper/view/14525}
  {Improving multi-document summarization via text classification}.
\newblock In \emph{Proceedings of the Thirty-First {AAAI} Conference on
  Artificial Intelligence, February 4-9, 2017, San Francisco, California,
  {USA}}, pages 3053--3059. {AAAI} Press.

\bibitem[{Cao et~al.(2016)Cao, Li, Li, Wei, and Li}]{cao-etal-2016-attsum}
Ziqiang Cao, Wenjie Li, Sujian Li, Furu Wei, and Yanran Li. 2016.
\newblock \href {https://www.aclweb.org/anthology/C16-1053/} {Attsum: Joint
  learning of focusing and summarization with neural attention}.
\newblock In \emph{{COLING} 2016, 26th International Conference on
  Computational Linguistics, Proceedings of the Conference: Technical Papers,
  December 11-16, 2016, Osaka, Japan}, pages 547--556. {ACL}.

\bibitem[{Carbonell and Goldstein(1998)}]{carbonell1998use}
Jaime~G Carbonell and Jade Goldstein. 1998.
\newblock The use of mmr and diversity-based reranking for reodering documents
  and producing summaries.

\bibitem[{Chen and Bansal(2018)}]{chen-bansal-2018-fast}
Yen{-}Chun Chen and Mohit Bansal. 2018.
\newblock \href {https://doi.org/10.18653/v1/P18-1063} {Fast abstractive
  summarization with reinforce-selected sentence rewriting}.
\newblock In \emph{Proceedings of the 56th Annual Meeting of the Association
  for Computational Linguistics, {ACL} 2018, Melbourne, Australia, July 15-20,
  2018, Volume 1: Long Papers}, pages 675--686. Association for Computational
  Linguistics.

\bibitem[{Cho et~al.(2019)Cho, Lebanoff, Foroosh, and
  Liu}]{cho-etal-2019-improving}
Sangwoo Cho, Logan Lebanoff, Hassan Foroosh, and Fei Liu. 2019.
\newblock \href {https://doi.org/10.18653/v1/p19-1098} {Improving the
  similarity measure of determinantal point processes for extractive
  multi-document summarization}.
\newblock In \emph{Proceedings of the 57th Conference of the Association for
  Computational Linguistics, {ACL} 2019, Florence, Italy, July 28- August 2,
  2019, Volume 1: Long Papers}, pages 1027--1038. Association for Computational
  Linguistics.

\bibitem[{Devlin et~al.(2019)Devlin, Chang, Lee, and
  Toutanova}]{devlin2018bert}
Jacob Devlin, Ming{-}Wei Chang, Kenton Lee, and Kristina Toutanova. 2019.
\newblock \href {https://doi.org/10.18653/v1/n19-1423} {{BERT:} pre-training of
  deep bidirectional transformers for language understanding}.
\newblock In \emph{Proceedings of the 2019 Conference of the North American
  Chapter of the Association for Computational Linguistics: Human Language
  Technologies, {NAACL-HLT} 2019, Minneapolis, MN, USA, June 2-7, 2019, Volume
  1 (Long and Short Papers)}, pages 4171--4186. Association for Computational
  Linguistics.

\bibitem[{Erkan and Radev(2004)}]{erkan2004lexrank}
G{\"{u}}nes Erkan and Dragomir~R. Radev. 2004.
\newblock \href {https://doi.org/10.1613/jair.1523} {Lexrank: Graph-based
  lexical centrality as salience in text summarization}.
\newblock \emph{J. Artif. Intell. Res.}, 22:457--479.

\bibitem[{Ganesan et~al.(2010)Ganesan, Zhai, and
  Han}]{ganesan-etal-2010-opinosis}
Kavita Ganesan, ChengXiang Zhai, and Jiawei Han. 2010.
\newblock \href {https://www.aclweb.org/anthology/C10-1039/} {Opinosis: {A}
  graph based approach to abstractive summarization of highly redundant
  opinions}.
\newblock In \emph{{COLING} 2010, 23rd International Conference on
  Computational Linguistics, Proceedings of the Conference, 23-27 August 2010,
  Beijing, China}, pages 340--348. Tsinghua University Press.

\bibitem[{Gao et~al.(2019)Gao, Meyer, Mesgar, and Gurevych}]{gao2019reward}
Yang Gao, Christian~M. Meyer, Mohsen Mesgar, and Iryna Gurevych. 2019.
\newblock \href {https://doi.org/10.24963/ijcai.2019/326} {Reward learning for
  efficient reinforcement learning in extractive document summarisation}.
\newblock In \emph{Proceedings of the Twenty-Eighth International Joint
  Conference on Artificial Intelligence, {IJCAI} 2019, Macao, China, August
  10-16, 2019}, pages 2350--2356. ijcai.org.

\bibitem[{Gillick and Favre(2009)}]{gillick-favre-2009-scalable}
Dan Gillick and Benoit Favre. 2009.
\newblock \href {https://www.aclweb.org/anthology/W09-1802} {A scalable global
  model for summarization}.
\newblock In \emph{Proceedings of the Workshop on Integer Linear Programming
  for Natural Language Processing}, pages 10--18, Boulder, Colorado.
  Association for Computational Linguistics.

\bibitem[{Graff et~al.(2003)Graff, Kong, Chen, and Maeda}]{graff2003english}
David Graff, Junbo Kong, Ke~Chen, and Kazuaki Maeda. 2003.
\newblock English gigaword.
\newblock \emph{Linguistic Data Consortium, Philadelphia}, 4(1):34.

\bibitem[{Gu et~al.(2020)Gu, Mao, Han, Liu, Wu, Yu, Finnie, Yu, Zhai, and
  Zukoski}]{gu2020generating}
Xiaotao Gu, Yuning Mao, Jiawei Han, Jialu Liu, You Wu, Cong Yu, Daniel Finnie,
  Hongkun Yu, Jiaqi Zhai, and Nicholas Zukoski. 2020.
\newblock \href {https://doi.org/10.1145/3366423.3380247} {Generating
  representative headlines for news stories}.
\newblock In \emph{{WWW} '20: The Web Conference 2020, Taipei, Taiwan, April
  20-24, 2020}, pages 1773--1784. {ACM} / {IW3C2}.

\bibitem[{Haghighi and Vanderwende(2009)}]{haghighi-vanderwende-2009-exploring}
Aria Haghighi and Lucy Vanderwende. 2009.
\newblock \href {https://www.aclweb.org/anthology/N09-1041/} {Exploring content
  models for multi-document summarization}.
\newblock In \emph{Human Language Technologies: Conference of the North
  American Chapter of the Association of Computational Linguistics,
  Proceedings, May 31 - June 5, 2009, Boulder, Colorado, {USA}}, pages
  362--370. The Association for Computational Linguistics.

\bibitem[{Haveliwala(2002)}]{haveliwala2002topic}
Taher~H. Haveliwala. 2002.
\newblock \href {https://doi.org/10.1145/511446.511513} {Topic-sensitive
  pagerank}.
\newblock In \emph{Proceedings of the Eleventh International World Wide Web
  Conference, {WWW} 2002, May 7-11, 2002, Honolulu, Hawaii, {USA}}, pages
  517--526. {ACM}.

\bibitem[{Hinton et~al.(2018)Hinton, Sabour, and Frosst}]{hinton2018matrix}
Geoffrey~E. Hinton, Sara Sabour, and Nicholas Frosst. 2018.
\newblock \href {https://openreview.net/forum?id=HJWLfGWRb} {Matrix capsules
  with {EM} routing}.
\newblock In \emph{6th International Conference on Learning Representations,
  {ICLR} 2018, Vancouver, BC, Canada, April 30 - May 3, 2018, Conference Track
  Proceedings}. OpenReview.net.

\bibitem[{Hong et~al.(2014)Hong, Conroy, Favre, Kulesza, Lin, and
  Nenkova}]{hong-etal-2014-repository}
Kai Hong, John~M. Conroy, Beno{\^{\i}}t Favre, Alex Kulesza, Hui Lin, and Ani
  Nenkova. 2014.
\newblock \href
  {http://www.lrec-conf.org/proceedings/lrec2014/summaries/1093.html} {A
  repository of state of the art and competitive baseline summaries for generic
  news summarization}.
\newblock In \emph{Proceedings of the Ninth International Conference on
  Language Resources and Evaluation, {LREC} 2014, Reykjavik, Iceland, May
  26-31, 2014}, pages 1608--1616. European Language Resources Association
  {(ELRA)}.

\bibitem[{Huang et~al.(2015)Huang, Xu, and Yu}]{huang2015bidirectional}
Zhiheng Huang, Wei Xu, and Kai Yu. 2015.
\newblock \href {http://arxiv.org/abs/1508.01991} {Bidirectional {LSTM-CRF}
  models for sequence tagging}.
\newblock \emph{CoRR}, abs/1508.01991.

\bibitem[{Kim(2014)}]{kim-2014-convolutional}
Yoon Kim. 2014.
\newblock \href {https://doi.org/10.3115/v1/d14-1181} {Convolutional neural
  networks for sentence classification}.
\newblock In \emph{Proceedings of the 2014 Conference on Empirical Methods in
  Natural Language Processing, {EMNLP} 2014, October 25-29, 2014, Doha, Qatar,
  {A} meeting of SIGDAT, a Special Interest Group of the {ACL}}, pages
  1746--1751. {ACL}.

\bibitem[{Kulesza and Taskar(2012)}]{kulesza2012determinantal}
Alex Kulesza and Ben Taskar. 2012.
\newblock \href {https://doi.org/10.1561/2200000044} {Determinantal point
  processes for machine learning}.
\newblock \emph{Foundations and Trends in Machine Learning}, 5(2-3):123--286.

\bibitem[{Latkowski(2018)}]{siamese-github}
Tomasz Latkowski. 2018.
\newblock Siamese deep neural networks for semantic similarity.
\newblock \url{https://github.com/tlatkowski/multihead-siamese-nets}.

\bibitem[{Lebanoff et~al.(2018)Lebanoff, Song, and
  Liu}]{lebanoff-etal-2018-adapting}
Logan Lebanoff, Kaiqiang Song, and Fei Liu. 2018.
\newblock \href {https://doi.org/10.18653/v1/d18-1446} {Adapting the neural
  encoder-decoder framework from single to multi-document summarization}.
\newblock In \emph{Proceedings of the 2018 Conference on Empirical Methods in
  Natural Language Processing, Brussels, Belgium, October 31 - November 4,
  2018}, pages 4131--4141. Association for Computational Linguistics.

\bibitem[{Lewis et~al.(2019)Lewis, Liu, Goyal, Ghazvininejad, Mohamed, Levy,
  Stoyanov, and Zettlemoyer}]{lewis2019bart}
Mike Lewis, Yinhan Liu, Naman Goyal, Marjan Ghazvininejad, Abdelrahman Mohamed,
  Omer Levy, Veselin Stoyanov, and Luke Zettlemoyer. 2019.
\newblock \href {http://arxiv.org/abs/1910.13461} {{BART:} denoising
  sequence-to-sequence pre-training for natural language generation,
  translation, and comprehension}.
\newblock \emph{CoRR}, abs/1910.13461.

\bibitem[{Li et~al.(2017)Li, Wang, Lam, Ren, and Bing}]{li2017salience}
Piji Li, Zihao Wang, Wai Lam, Zhaochun Ren, and Lidong Bing. 2017.
\newblock \href {http://aaai.org/ocs/index.php/AAAI/AAAI17/paper/view/14613}
  {Salience estimation via variational auto-encoders for multi-document
  summarization}.
\newblock In \emph{Proceedings of the Thirty-First {AAAI} Conference on
  Artificial Intelligence, February 4-9, 2017, San Francisco, California,
  {USA}}, pages 3497--3503. {AAAI} Press.

\bibitem[{Lin(2004)}]{lin-2004-rouge}
Chin-Yew Lin. 2004.
\newblock {ROUGE}: A package for automatic evaluation of summaries.
\newblock In \emph{Text Summarization Branches Out}.

\bibitem[{Liu and Lapata(2019)}]{liu-lapata-2019-text}
Yang Liu and Mirella Lapata. 2019.
\newblock \href {https://doi.org/10.18653/v1/D19-1387} {Text summarization with
  pretrained encoders}.
\newblock In \emph{Proceedings of the 2019 Conference on Empirical Methods in
  Natural Language Processing and the 9th International Joint Conference on
  Natural Language Processing, {EMNLP-IJCNLP} 2019, Hong Kong, China, November
  3-7, 2019}, pages 3728--3738. Association for Computational Linguistics.

\bibitem[{Mao et~al.(2020)Mao, Liu, Zhu, Ren, and Han}]{mao-etal-2020-facet}
Yuning Mao, Liyuan Liu, Qi~Zhu, Xiang Ren, and Jiawei Han. 2020.
\newblock \href {https://doi.org/10.18653/v1/2020.acl-main.445} {Facet-aware
  evaluation for extractive summarization}.
\newblock In \emph{Proceedings of the 58th Annual Meeting of the Association
  for Computational Linguistics}, pages 4941--4957, Online. Association for
  Computational Linguistics.

\bibitem[{Mao et~al.(2018)Mao, Ren, Shen, Gu, and Han}]{mao-etal-2018-end}
Yuning Mao, Xiang Ren, Jiaming Shen, Xiaotao Gu, and Jiawei Han. 2018.
\newblock \href {https://doi.org/10.18653/v1/P18-1229} {End-to-end
  reinforcement learning for automatic taxonomy induction}.
\newblock In \emph{Proceedings of the 56th Annual Meeting of the Association
  for Computational Linguistics, {ACL} 2018, Melbourne, Australia, July 15-20,
  2018, Volume 1: Long Papers}, pages 2462--2472. Association for Computational
  Linguistics.

\bibitem[{Mao et~al.(2019)Mao, Tian, Han, and Ren}]{DBLP:conf/emnlp/MaoTHR19}
Yuning Mao, Jingjing Tian, Jiawei Han, and Xiang Ren. 2019.
\newblock \href {https://doi.org/10.18653/v1/D19-1042} {Hierarchical text
  classification with reinforced label assignment}.
\newblock In \emph{Proceedings of the 2019 Conference on Empirical Methods in
  Natural Language Processing and the 9th International Joint Conference on
  Natural Language Processing, {EMNLP-IJCNLP} 2019, Hong Kong, China, November
  3-7, 2019}, pages 445--455. Association for Computational Linguistics.

\bibitem[{Nallapati et~al.(2017)Nallapati, Zhai, and
  Zhou}]{nallapati2017summarunner}
Ramesh Nallapati, Feifei Zhai, and Bowen Zhou. 2017.
\newblock \href {http://aaai.org/ocs/index.php/AAAI/AAAI17/paper/view/14636}
  {Summarunner: {A} recurrent neural network based sequence model for
  extractive summarization of documents}.
\newblock In \emph{Proceedings of the Thirty-First {AAAI} Conference on
  Artificial Intelligence, February 4-9, 2017, San Francisco, California,
  {USA}}, pages 3075--3081. {AAAI} Press.

\bibitem[{Nallapati et~al.(2016)Nallapati, Zhou, dos Santos,
  G{\"{u}}l{\c{c}}ehre, and Xiang}]{nallapati-etal-2016-abstractive}
Ramesh Nallapati, Bowen Zhou, C{\'{\i}}cero~Nogueira dos Santos, {\c{C}}aglar
  G{\"{u}}l{\c{c}}ehre, and Bing Xiang. 2016.
\newblock \href {https://doi.org/10.18653/v1/k16-1028} {Abstractive text
  summarization using sequence-to-sequence rnns and beyond}.
\newblock In \emph{Proceedings of the 20th {SIGNLL} Conference on Computational
  Natural Language Learning, CoNLL 2016, Berlin, Germany, August 11-12, 2016},
  pages 280--290. {ACL}.

\bibitem[{Narayan et~al.(2018)Narayan, Cohen, and
  Lapata}]{narayan-etal-2018-ranking}
Shashi Narayan, Shay~B. Cohen, and Mirella Lapata. 2018.
\newblock \href {https://doi.org/10.18653/v1/n18-1158} {Ranking sentences for
  extractive summarization with reinforcement learning}.
\newblock In \emph{Proceedings of the 2018 Conference of the North American
  Chapter of the Association for Computational Linguistics: Human Language
  Technologies, {NAACL-HLT} 2018, New Orleans, Louisiana, USA, June 1-6, 2018,
  Volume 1 (Long Papers)}, pages 1747--1759. Association for Computational
  Linguistics.

\bibitem[{Nayeem et~al.(2018)Nayeem, Fuad, and
  Chali}]{nayeem-etal-2018-abstractive}
Mir~Tafseer Nayeem, Tanvir~Ahmed Fuad, and Yllias Chali. 2018.
\newblock \href {https://www.aclweb.org/anthology/C18-1102/} {Abstractive
  unsupervised multi-document summarization using paraphrastic sentence
  fusion}.
\newblock In \emph{Proceedings of the 27th International Conference on
  Computational Linguistics, {COLING} 2018, Santa Fe, New Mexico, USA, August
  20-26, 2018}, pages 1191--1204. Association for Computational Linguistics.

\bibitem[{Owczarzak and Dang(2011)}]{owczarzak2011overview}
Karolina Owczarzak and Hoa~Trang Dang. 2011.
\newblock Overview of the tac 2011 summarization track: Guided task and aesop
  task.
\newblock In \emph{Proceedings of the Text Analysis Conference (TAC 2011),
  Gaithersburg, Maryland, USA, November}.

\bibitem[{Paul and James(2004)}]{paul2004introduction}
Over Paul and Yen James. 2004.
\newblock An introduction to duc-2004.
\newblock In \emph{DUC}.

\bibitem[{Paulus et~al.(2018)Paulus, Xiong, and Socher}]{paulus2017deep}
Romain Paulus, Caiming Xiong, and Richard Socher. 2018.
\newblock \href {https://openreview.net/forum?id=HkAClQgA-} {A deep reinforced
  model for abstractive summarization}.
\newblock In \emph{6th International Conference on Learning Representations,
  {ICLR} 2018, Vancouver, BC, Canada, April 30 - May 3, 2018, Conference Track
  Proceedings}. OpenReview.net.

\bibitem[{See et~al.(2017)See, Liu, and Manning}]{see-etal-2017-get}
Abigail See, Peter~J. Liu, and Christopher~D. Manning. 2017.
\newblock \href {https://doi.org/10.18653/v1/P17-1099} {Get to the point:
  Summarization with pointer-generator networks}.
\newblock In \emph{Proceedings of the 55th Annual Meeting of the Association
  for Computational Linguistics, {ACL} 2017, Vancouver, Canada, July 30 -
  August 4, Volume 1: Long Papers}, pages 1073--1083. Association for
  Computational Linguistics.

\bibitem[{Song et~al.(2018)Song, Zhao, and Liu}]{song-etal-2018-structure}
Kaiqiang Song, Lin Zhao, and Fei Liu. 2018.
\newblock \href {https://www.aclweb.org/anthology/C18-1146/} {Structure-infused
  copy mechanisms for abstractive summarization}.
\newblock In \emph{Proceedings of the 27th International Conference on
  Computational Linguistics, {COLING} 2018, Santa Fe, New Mexico, USA, August
  20-26, 2018}, pages 1717--1729. Association for Computational Linguistics.

\bibitem[{Tan et~al.(2017)Tan, Wan, and Xiao}]{tan-etal-2017-abstractive}
Jiwei Tan, Xiaojun Wan, and Jianguo Xiao. 2017.
\newblock \href {https://doi.org/10.18653/v1/P17-1108} {Abstractive document
  summarization with a graph-based attentional neural model}.
\newblock In \emph{Proceedings of the 55th Annual Meeting of the Association
  for Computational Linguistics, {ACL} 2017, Vancouver, Canada, July 30 -
  August 4, Volume 1: Long Papers}, pages 1171--1181. Association for
  Computational Linguistics.

\bibitem[{Vanderwende et~al.(2007)Vanderwende, Suzuki, Brockett, and
  Nenkova}]{vanderwende2007beyond}
Lucy Vanderwende, Hisami Suzuki, Chris Brockett, and Ani Nenkova. 2007.
\newblock \href {https://doi.org/10.1016/j.ipm.2007.01.023} {Beyond sumbasic:
  Task-focused summarization with sentence simplification and lexical
  expansion}.
\newblock \emph{Inf. Process. Manag.}, 43(6):1606--1618.

\bibitem[{Vinyals et~al.(2016)Vinyals, Bengio, and Kudlur}]{vinyals2015order}
Oriol Vinyals, Samy Bengio, and Manjunath Kudlur. 2016.
\newblock \href {http://arxiv.org/abs/1511.06391} {Order matters: Sequence to
  sequence for sets}.
\newblock In \emph{4th International Conference on Learning Representations,
  {ICLR} 2016, San Juan, Puerto Rico, May 2-4, 2016, Conference Track
  Proceedings}.

\bibitem[{Wang et~al.(2017)Wang, Liu, Sui, and Chang}]{wang-etal-2017-affinity}
Kexiang Wang, Tianyu Liu, Zhifang Sui, and Baobao Chang. 2017.
\newblock \href {https://doi.org/10.18653/v1/d17-1020} {Affinity-preserving
  random walk for multi-document summarization}.
\newblock In \emph{Proceedings of the 2017 Conference on Empirical Methods in
  Natural Language Processing, {EMNLP} 2017, Copenhagen, Denmark, September
  9-11, 2017}, pages 210--220. Association for Computational Linguistics.

\bibitem[{Williams(1992)}]{williams1992simple}
Ronald~J. Williams. 1992.
\newblock \href {https://doi.org/10.1007/BF00992696} {Simple statistical
  gradient-following algorithms for connectionist reinforcement learning}.
\newblock \emph{Mach. Learn.}, 8:229--256.

\bibitem[{Wong et~al.(2008)Wong, Wu, and Li}]{wong-etal-2008-extractive}
Kam-Fai Wong, Mingli Wu, and Wenjie Li. 2008.
\newblock \href {https://www.aclweb.org/anthology/C08-1124} {Extractive
  summarization using supervised and semi-supervised learning}.
\newblock In \emph{Proceedings of the 22nd International Conference on
  Computational Linguistics (Coling 2008)}, pages 985--992, Manchester, UK.
  Coling 2008 Organizing Committee.

\bibitem[{Yasunaga et~al.(2017)Yasunaga, Zhang, Meelu, Pareek, Srinivasan, and
  Radev}]{yasunaga-etal-2017-graph}
Michihiro Yasunaga, Rui Zhang, Kshitijh Meelu, Ayush Pareek, Krishnan
  Srinivasan, and Dragomir~R. Radev. 2017.
\newblock \href {https://doi.org/10.18653/v1/K17-1045} {Graph-based neural
  multi-document summarization}.
\newblock In \emph{Proceedings of the 21st Conference on Computational Natural
  Language Learning (CoNLL 2017), Vancouver, Canada, August 3-4, 2017}, pages
  452--462. Association for Computational Linguistics.

\bibitem[{Zhang et~al.(2018)Zhang, Tan, and Wan}]{zhang-etal-2018-adapting}
Jianmin Zhang, Jiwei Tan, and Xiaojun Wan. 2018.
\newblock \href {https://doi.org/10.18653/v1/w18-6545} {Adapting neural
  single-document summarization model for abstractive multi-document
  summarization: {A} pilot study}.
\newblock In \emph{Proceedings of the 11th International Conference on Natural
  Language Generation, Tilburg University, The Netherlands, November 5-8,
  2018}, pages 381--390. Association for Computational Linguistics.

\bibitem[{Zhang et~al.(2020)Zhang, Zhao, Saleh, and Liu}]{zhang2019pegasus}
Jingqing Zhang, Yao Zhao, Mohammad Saleh, and Peter~J. Liu. 2020.
\newblock \href {http://arxiv.org/abs/1912.08777} {{PEGASUS:} pre-training with
  extracted gap-sentences for abstractive summarization}.
\newblock \emph{ICML}, abs/1912.08777.

\bibitem[{Zhao et~al.(2019)Zhao, Peyrard, Liu, Gao, Meyer, and
  Eger}]{zhao-etal-2019-moverscore}
Wei Zhao, Maxime Peyrard, Fei Liu, Yang Gao, Christian~M. Meyer, and Steffen
  Eger. 2019.
\newblock \href {https://doi.org/10.18653/v1/D19-1053} {Moverscore: Text
  generation evaluating with contextualized embeddings and earth mover
  distance}.
\newblock In \emph{Proceedings of the 2019 Conference on Empirical Methods in
  Natural Language Processing and the 9th International Joint Conference on
  Natural Language Processing, {EMNLP-IJCNLP} 2019, Hong Kong, China, November
  3-7, 2019}, pages 563--578. Association for Computational Linguistics.

\end{thebibliography}
\bibliographystyle{acl_natbib}

\newpage
\clearpage
\appendix

\section{Experimental Details}
\label{app:baseline}

\subsection{Dataset Statistics}\label{app:data}
We list in Table~\ref{table:dataset} the details of datasets used in our experiments.
SDS methods usually take the first 256 or 512 words in the document as model input, which is infeasible for the input size of MDS (5,000 to 7,000 words on average).
\begin{table}[h]
    \centering

    \resizebox{\columnwidth}{!}{
    \begin{tabular}{l rrrrrr}
        \toprule
         \textbf{Dataset} & \#$\gD$ & $\sum |\gD|$ & $\overline{\sum |D_i|}$ & $\min \sum |D_i|$ & $\max \sum |D_i|$  & $\overline{\sum |s_j|}$\\
        \midrule
        DUC 2003  & 30 & 298 & 259.0 & 98 & 502  &6830.5\\
        DUC 2004  & 50 & 500 & 265.4 & 152 & 605 &6987.1\\
        TAC 2008-2010 & 138 & 1380 & 236.9 & 41 & 649  & 5978.4\\
        TAC 2011 & 44 & 440 & 204.9 & 48 & 486 &5146.0\\
        \bottomrule
    \end{tabular}
    }
    
    \upv
        \caption{\textbf{Dataset statistics}. \#$\gD$ and $\sum |\gD|$ denote the number of document sets and the number of documents in total. $\overline{\sum |D_i|}$, $\min \sum |D_i|$, and $\max \sum |D_i|$ denote the average / min / max number of sentences in a document set. $\overline{\sum |s_j|}$ denotes the average number of words in a document set.}
        \label{table:dataset}
    \downv
\end{table}

\subsection{Remarks on Experimental Setup}
We note that there are plenty of inconsistencies in the previous work on MDS and some results cannot be directly compared with ours.
Specifically, there are three major differences that may lead to incomparable results as follows.
First, while in the original DUC competitions an output length of 665 bytes is adopted, more recent studies mostly take a length limit of 100 words following \citet{hong-etal-2014-repository}, and some do not have any length limit (usually resulting in higher numbers).
Second, some papers report ROUGE recall \cite{yasunaga-etal-2017-graph,wang-etal-2017-affinity,cao2017improving,nayeem-etal-2018-abstractive,gao2019reward} while others (including ours) report ROUGE F1 following the trend on SDS \cite{lebanoff-etal-2018-adapting,zhang-etal-2018-adapting,cho-etal-2019-improving}.
Third, while DUC 2004 and TAC 2011 are usually used as test sets, the training sets used in different studies often vary. We follow the same setup as the compared methods to ensure a fair comparison.

\subsection{Description of Extractive Baselines}
SumBasic~\cite{vanderwende2007beyond} is based on word frequency and hypothesizes that the words occurring frequently are likely to be included in the summary. {KLSumm}~\cite{haghighi-vanderwende-2009-exploring} greedily extracts sentences as long as they can lead to a decrease in KL divergence.
{LexRank}~\cite{erkan2004lexrank} computes sentence salience based on eigenvector centrality in a graph-based representation.
{Centroid}~\cite{hong-etal-2014-repository} measures sentence salience based on its cosine similarity with the document centroid, which is similar to the salience measure in MMR.
{ICSISumm}~\cite{gillick-favre-2009-scalable} uses integer linear programming (ILP) to extract a globally optimal set of sentences that can cover the most important concepts in the document set.
{DPP}~\cite{kulesza2012determinantal} handles sentence salience and redundancy through the determinantal point processes, in which many handcrafted features such as sentence length, sentence position, and personal pronouns are used.
DPP-Caps-Comb~\cite{cho-etal-2019-improving} improves upon DPP~\cite{kulesza2012determinantal} by replacing or combining the existing sentence salience and redundancy measures with capsule networks~\cite{hinton2018matrix}.
{rnn-ext + RL}~\cite{chen-bansal-2018-fast} is the SDS method that we base our work on. It is pre-trained on the CNN/Daily Mail SDS dataset~\cite{nallapati-etal-2016-abstractive}, and we test its performance with or without fine-tuning on the MDS training set.
The pre-trained abstractor in rnn-ext + RL is not used as we found it consistently leads to worse performance.

\subsection{Description of Abstractive Baselines}
{Opinosis}~\cite{ganesan-etal-2010-opinosis} generates summaries by finding salient paths on a word co-occurrence graph of the documents. 
{Extract+Rewrite}~\cite{song-etal-2018-structure} scores sentences by LexRank~\cite{erkan2004lexrank} and employs an encoder-decoder model pre-trained on Gigaword~\cite{graff2003english} to generate a title-like summary for each sentence.
{PG}~\cite{see-etal-2017-get} is one typical abstractive summarization method for SDS that conducts sequence-to-sequence learning with copy mechanism.
{PG-MMR}~\cite{lebanoff-etal-2018-adapting} adapts PG~\cite{see-etal-2017-get} to MDS by concatenating all of the documents in one document set and running pre-trained PG under the constraints of MMR on its vocabulary.

\label{sec:implementation}

\subsection{Implementation Details}
Following common practice, we only consider extracting sentences with reasonable length (\ie, 8 to 55 words)~\cite{erkan2004lexrank,yasunaga-etal-2017-graph}.
We filter sentences that start with a quotation mark or do not end with a period~\cite{wong-etal-2008-extractive,lebanoff-etal-2018-adapting}.
For MMR, we set $\lambda=0.6$ following~\citet{lebanoff-etal-2018-adapting}.
By default, we use TF-IDF features and cosine similarity for both sentence salience and redundancy measurement in MMR.
We prefer such measurements instead of ROUGE-based measures~\cite{lebanoff-etal-2018-adapting} and advanced neural-based measures~\cite{cho-etal-2019-improving,devlin2018bert,siamese-github} as they are faster to compute and comparable in performance.
We pre-train {rnn-ext + RL}~\cite{chen-bansal-2018-fast} on the CNN/Daily Mail SDS dataset~\cite{nallapati-etal-2016-abstractive} as in~\citet{lebanoff-etal-2018-adapting} but continue fine-tuning on the in-domain training set.
We train \ours using an Adam optimizer with learning rate 5e-4 for \oursSoftAttn and 1e-3 for the other variants without weight decay.
we tested various reward functions, such as different ROUGE metrics, the MMR scores, and intrinsic measures based on sentence representation, and found them comparable or worse than the current one. One may also use other semantic metrics such as MoverScore~\cite{zhao-etal-2019-moverscore} and FAR~\cite{mao-etal-2020-facet}.

\section{Detailed Analysis of \ours}
\label{sec:details}

\start{Additional Illustration}
\begin{figure}[t]
    \includegraphics[width=.99\linewidth]{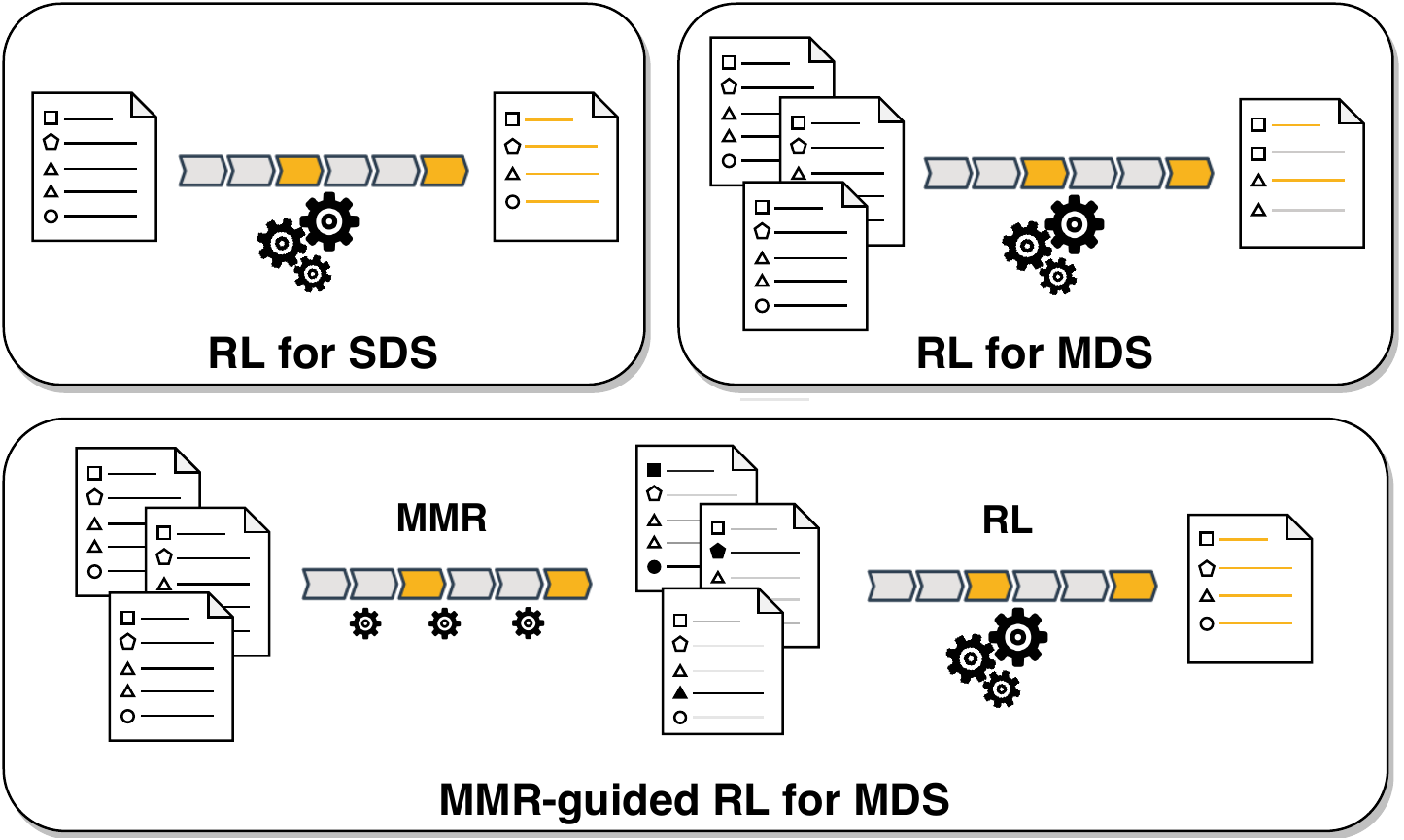}
    \upv
    \caption{We use the same shape to denote semantically similar sentences. Directly applying RL to MDS encounters large search space and high redundancy, resulting in repeated summaries. MMR guides RL by attending to salient and non-redundant candidates.}
    \label{fig:teaser}
    \downv
\end{figure}
We provide an illustration in Fig.~\ref{fig:teaser} to better elucidate the motivation of \ours.
Note that \ours is mostly based on SDS architectures while achieving state-of-the-art performance on MDS, while existing combination approaches that achieve decent performance (\eg, DPP-Caps) are based on MDS architectures.

\start{Runtime and Memory Usage}
\ours is time and space efficient for two reasons.
First, its hierarchical sentence encoding is much faster than a word-level sequence encoding mechanism while still capturing global context.
Second, the guidance of MMR provides \ours with a ``warmup'' effect, leading to faster convergence.
In our experiments, one epoch of \ours takes 0.87 to 0.91s on a GTX 1080 GPU with less than 1.2 GB memory usage. 
The number of epochs is set to 10,000 and we adopt early stopping -- the training process terminates if \ours cannot achieve better results on the validation set after 30 continuous evaluations.
As a result, the runs often terminate before 5,000 epochs, and the overall training time ranges from 40 to 90 minutes.

\begin{table*}[ht]
    \footnotesize

        \scalebox{1}{
        \begin{tabular}{|p{2\columnwidth}|}
            \toprule
             \textbf{RL}: 1. President Clinton made an unusual, direct appeal to North Korea on 
Saturday to set aside any nuclear ambitions in favor of strengthening 
ties to South Korea and the United States.\\
2. SEOUL, South Korea (AP)  U.S. President Bill Clinton won South Korea's 
support Saturday for confronting North Korea over a suspected nuclear 
site, and he warned the North's communist leaders not to squander 
\colorR{an historic chance to make a lasting peace} on the peninsula.\\
3. SEOUL, South Korea (AP)  U.S. President Bill Clinton won South Korea's 
support Saturday for confronting North Korea over a suspected nuclear 
site, and he warned the North's communist leaders not to squander 
\colorR{a chance to achieve lasting peace} on the peninsula.\\
             
             \medskip
             \textbf{\ours}: 1. SEOUL, South Korea (AP)  U.S. President Bill Clinton won South Korea's 
support ... an historic chance to make a lasting peace on the peninsula.\\
2. The North Koreans have denied that the 
complex, which is being built on a mountainside about 25 miles northeast 
of Yongbyon, the former North Korean nuclear research center, is intended 
to be used for a nuclear weapons program.\\
3. The United States and North Korea are set to 
resume talks Friday about inspections of an underground North Korean 
site suspected of being used to produce nuclear weapons.\\
             \midrule
             
             \midrule
             \textbf{RL}: 1. Galina Starovoitova, 52, 
a leader of the liberal Russia's Democratic Choice party, \colorR{was shot 
dead} by unidentified assailants on the stairs of her apartment building 
in St. Petersburg on Friday night.\\
2. A liberal lawmaker who planned to run for president in Russia's next 
elections \colorR{was shot to death} Friday in St. Petersburg, \colorR{police said}.\\
3. A liberal lawmaker who planned to run for president in Russia's next 
elections \colorR{was shot to death} Friday in St. Petersburg, \colorR{police said}.\\
4. A liberal lawmaker who planned to run for president in Russia's next 
elections \colorR{was killed} Friday in St. Petersburg, \colorR{a news report said}.\\

             \medskip
             \textbf{\ours}: 1. Galina Starovoitova, 52, 
a leader of the liberal Russia's Democratic Choice party, was shot 
dead by unidentified assailants on the stairs of her apartment...\\
2. Starovoitova tried to run for president in 
the 1996 elections but her registration was turned down for technical 
reasons.\\
3. Like that fictional 
crime, which shone a light on social ferment in the St. Petersburg 
of its day, the death of Starovoitova was immediately seized upon 
as a seminal event in the Russia of the late 1990s.\\
4. She was a member of the Russian parliament and 
a recently declared candidate for governor of the region around St. 
Petersburg.\\
            \bottomrule
        \end{tabular}
        }
        
                \upv
                \caption{\textbf{Case studies reveal the insufficient redundancy measure in vanilla RL}. Note that the 2nd and 3rd extracted sentences of RL in the second example  are the same but from different documents, which is quite typical in news reports.}        \label{tab:redundancy}
                \downv
    \end{table*}
\start{Detailed Examples}
In Table~\ref{tab:redundancy}, we show the extracted summaries of vanilla RL and \ours for the same document set.
Without the guidance of MMR, the RL agent is much more likely to extract redundant sentences.
In the first example, RL extracts two semantically equivalent sentences from two different documents. These two sentences would have similar sentence representation $\vh^i_j$, and the latent state representation $\vg_t$ itself might not be enough to avoid redundant extraction.
In contrast, \ours selects diverse sentences after extracting the same original sentence as RL thanks to the explicit redundancy measure in MMR.
In the second example, the issue of redundancy in RL is even more severe -- all four extracted sentences of RL are covering the same aspect of the news.
\ours again balances sentence salience and redundancy better than vanilla RL, favoring diverse sentences.
Such results imply that pure neural representation is insufficient for redundancy avoidance in MDS and that classical approaches can serve as a complement.

\end{document}